# Bayesian Learning in Undirected Graphical Models: Approximate MCMC algorithms

**Iain Murray** and **Zoubin Ghahramani**
Gatsby Computational Neuroscience Unit
University College London, London WC1N 3AR, UK
http://www.gatsby.ucl.ac.uk/
{i.murray,zoubin}@gatsby.ucl.ac.uk

## Abstract

Bayesian learning in undirected graphical models—computing posterior distributions over parameters and predictive quantities—is exceptionally difficult. We conjecture that for general undirected models, there are no tractable MCMC (Markov Chain Monte Carlo) schemes giving the correct equilibrium distribution over parameters. While this intractability, due to the partition function, is familiar to those performing parameter optimisation, Bayesian learning of posterior distributions over undirected model parameters has been unexplored and poses novel challenges. We propose several approximate MCMC schemes and test on fully observed binary models (Boltzmann machines) for a small coronary heart disease data set and larger artificial systems. While approximations must perform well on the model, their interaction with the sampling scheme is also important. Samplers based on variational mean-field approximations generally performed poorly, more advanced methods using loopy propagation, brief sampling and stochastic dynamics lead to acceptable parameter posteriors. Finally, we demonstrate these techniques on a Markov random field with hidden variables.

## 1 Introduction

Probabilistic graphical models are an elegant and powerful framework for representing distributions over many random variables. Undirected graphs provide a natural description of soft constraints between variables. Mutual compatibilities amongst variables, $\mathbf{x} = (x_1, \ldots x_k)$, are described by a factorised joint probability distribution:

$$p(\mathbf{x}|\theta) = \frac{1}{Z(\theta)} \exp \left\{ \sum_j \phi_j(\mathbf{x}_{C_j}, \theta_j) \right\}, \quad (1)$$

where $C_j \subset \{1, \ldots, k\}$ indexes a subset of the variables and $\phi_j$ is a potential function, parameterised by $\theta_j$, expressing compatibilities amongst $\mathbf{x}_{C_j}$. The *partition function* or normalisation constant

$$Z(\theta) = \sum_{\mathbf{x}} \exp \left\{ \sum_j \phi_j(\mathbf{x}_{C_j}, \theta_j) \right\} \quad (2)$$

is the (usually intractable) sum or integral over all configurations of the variables. The undirected model representing the conditional independencies implied by the factorization (1) has a node for each variable and an undirected edge connecting every pair of variables $x_i$—$x_\ell$, if $i, \ell \in C_j$ for some $j$. The subsets $C_j$ are therefore cliques (fully connected subgraphs) of the whole graph. An alternative and more general representation of undirected models is a *factor graph*. Factor graphs are bipartite graphs consisting of two types of nodes, one type representing the variables $i \in \{1, \ldots, k\}$ and the other type the factors $j$ appearing in the product (1). A variable node $i$ is connected via an undirected edge to a factor node $j$ if $i \in C_j$.

This work focuses on representing the parameter posterior $p(\theta|\mathbf{x})$ using samples, which can be used in approximating distributions over predictive quantities. Averaging over the parameter posterior can avoid the overfitting associated with optimisation. While sampling from parameters has attracted much attention, and is often tractable, in directed models, it is much more difficult for all but the most trivial[1] *undirected* graphical models. While directed models are a more natural tool for modelling causal relationships, the soft constraints provided by undirected models have proven

---

[1] i.e., low tree-width graphs, graphical Gaussian models and small contingency tables.

useful in a variety of problem domains; we briefly mention six applications.

**(a)** In computer vision [1] Markov random fields (MRFs), a form of undirected model, are used to model the soft constraint a pixel or image feature imposes on nearby pixels or features; this use of MRFs grew out of a long tradition in spatial statistics [2]. **(b)** In language modelling a common form of sentence model measures a large number of features of a sentence $f_j(s)$, such as the presence of a word, subject-verb agreement, the output of a parser on the sentence, etc, and assigns each such feature a weight $\lambda_j$. A random field model of this is then $p(s|\lambda) = (1/Z(\lambda)) \exp\{\sum_j \lambda_j f_j(s)\}$ where the weights can be learned via maximum likelihood iterative scaling methods [3]. **(c)** These undirected models can be extended to coreference analysis, which deals with determining, for example, whether two items (e.g., strings, citations) refer to the same underlying object [4]. **(d)** Undirected models have been used to model protein folding [5] and the soft constraints on the configuration of protein side chains [6]. **(e)** Semi-supervised classification is the problem of classifying a large number of unlabelled points using a small number of labelled points and some prior knowledge that nearby points have the same label. This problem can be approached by defining an undirected graphical model over both labelled and unlabelled data [7]. **(f)** Given a set of directed models $p(\mathbf{x}|\theta_j)$, the *products of experts* idea is a simple way of defining a more powerful (undirected) model by multiplying them: $p(\mathbf{x}|\theta) = (1/Z(\theta)) \prod_j p(\mathbf{x}|\theta_j)$ [8]. The product assigns high probability when there is consensus among component models.

Despite the long history and wide applicability of undirected models, surprisingly, Bayesian treatments of large undirected models are virtually non-existent! Indeed there *is* a related statistical literature on Bayesian inference in undirected models, log linear models, and contingency tables [9, 10, 11]. However, this literature assumes that the partition function $Z(\theta)$ can be computed exactly. But for all six machine learning applications of undirected models cited above, this assumption is unreasonable. This paper addresses Bayesian learning for models with intractable $Z(\theta)$.

We focus on a particularly simple and well-studied undirected model, the Boltzmann machine.

## 2  Bayesian Inference in Boltzmann Machines

A Boltzmann machine (BM) is a Markov random field which defines a probability distribution over a vector of binary variables $\mathbf{s} = [s_1, \ldots, s_k]$ where $s_i \in \{0, 1\}$:

$$p(\mathbf{s}|W) = \frac{1}{Z(W)} \exp\left\{\sum_{i<j} W_{ij} s_i s_j\right\} \quad (3)$$

The symmetric weight matrix $W$ parameterises this distribution. In a BM there are usually also linear bias terms $\sum_i b_i s_i$ in the exponent; we omit these biases to simplify notation, although the models in the experiments assume them. The undirected model for a BM has edges for all non-zero elements of $W$. Since the Boltzmann machine has only pairwise terms in the exponent, factor graphs provide a better representation for the model.

The usual algorithm for learning BMs is a maximum likelihood version of the EM algorithm (assuming some of the variables are hidden $\mathbf{s}_H$ and some observed $\mathbf{s}_O$) [12]. The gradient of the log probability is:

$$\frac{\partial \log p(\mathbf{s}|W)}{\partial W_{ij}} = \langle s_i s_j \rangle_c - \langle s_i s_j \rangle_u \quad (4)$$

where $\langle \cdot \rangle_c$ denotes expectation under the "clamped" data distribution $p(\mathbf{s}_H|\mathbf{s}_O, W)$ and $\langle \cdot \rangle_u$ denotes expectation under the "unclamped" distribution $p(\mathbf{s}|W)$. For a data set $\mathbf{S} = [\mathbf{s}^{(1)} \ldots \mathbf{s}^{(n)} \ldots \mathbf{s}^{(N)}]$ of i.i.d. data the gradient of the log likelihood is simply summed over $n$. For Boltzmann machines with large tree-width these expectations would take exponential time to compute, and the usual approach is to approximate them using Gibbs sampling or one of many more recent approximate inference algorithms.

Consider doing Bayesian inference for the parameters of a Boltzmann machine, i.e., computing $p(W|\mathbf{S})$. One can define a joint model:

$$p(W, \mathbf{S}) = \frac{1}{Z} \exp\left\{-\frac{1}{2\sigma^2} \sum_{j<i} W_{ij}^2 + \sum_n \sum_{j<i} W_{ij} s_i^{(n)} s_j^{(n)}\right\} \quad (5)$$

The first term acts like a prior, the normaliser $Z$ does not depend on $W$, and it is easy to see that $p(\mathbf{S}|W)$ is exactly the likelihood term for a Boltzmann machine with i.i.d. data: $p(\mathbf{S}|W) = \prod_n p(\mathbf{s}^{(n)}|W) = \prod_n (1/Z(W)) \exp\{\sum_{i<j} W_{ij} s_i^{(n)} s_j^{(n)}\}$. Moreover, it is very easy to sample from $p(W|\mathbf{S})$ since it is a multivariate Gaussian. Thus it appears that we have defined a joint distribution where the likelihood is exactly the BM model, and the posterior over parameters is trivial to sample from. Could Bayesian inference in Boltzmann machines be so simple?

Unfortunately, there is something deeply flawed with the above approach. By marginalisation of (5), the actual prior over the parameters must have been

$$p(W) = \sum_{\mathbf{S}} p(W, \mathbf{S}) \propto \mathcal{N}(0, \sigma^2 I) Z(W)^N. \quad (6)$$

However, this "prior" is dependent on the size of the data set! Moreover, the parametric form of the "prior" is very complicated, favouring weights with large partition functions—an effect that will overwhelm the Gaussian term. This is therefore not a valid hierarchical Bayesian model for a BM, and inferences from this model will be essentially meaningless.

The lesson from this simple example is the following: it is not possible to remove the partition function from the parameter posterior, as the "prior" that this would imply will be dependent on the number of data points. In order to have sensible parameter inferences, therefore, considering changes of the partition function with the parameters is unavoidable. Fortunately, there exist a large number of tools for approximating partition functions and their derivatives, given by expectations under (1). We now examine how approximations to partition functions and expectations can be used for approximate Bayesian inference in undirected models.

## 3 Monte Carlo Parameter Sampling

MCMC (Markov Chain Monte Carlo) methods allow us to draw correlated samples from a probability distribution with unknown normalisation. A rich set of methods are available [13], but as discussed above any scheme must compute a quantity related to the partition function before making any change to the parameters. We discuss two simple samplers that demonstrate the range of approximate methods available.

Consider the simplest Metropolis sampling scheme for the parameters of a Boltzmann machine given fully observed data. Starting from parameters $W$, assume that $W'$ is proposed from a symmetric proposal distribution $t(W'|W) = t(W|W')$. This proposal should be accepted with probability $a = \min(1, p(W'|\mathbf{S})/p(W|\mathbf{S}))$ where

$$\frac{p(W'|\mathbf{S})}{p(W|\mathbf{S})} = \frac{p(W')p(\mathbf{S}|W')}{p(W)p(\mathbf{S}|W)} \quad (7)$$

$$= \frac{p(W')}{p(W)} \left(\frac{Z(W)}{Z(W')}\right)^N \exp\left\{\sum_{n, i<j}(W'_{ij} - W_{ij}) s_i^{(n)} s_j^{(n)}\right\}.$$

For general BMs even a single step of this simple scheme is intractable due to $Z(W)$. One class of approach we will pursue is using deterministic tools to approximate $\tilde{Z}(W) \simeq Z(W)$ in the above expression. Clearly this results in an *approximate* sampler, which does not converge to the true equilibrium distribution over parameters. Moreover, it seems reckless to take an approximate quantity to the $N^{\text{th}}$ power. Despite these caveats we explore empirically whether approaches based on this class of approximation are viable.

Note that above we need only compute the ratio of the partition function at pairs of parameter settings, $Z(W)/Z(W')$. This ratio can be approximated directly by noting that:

$$\frac{Z(W)}{Z(W')} = \sum_{\mathbf{s}} e^{\{\sum_{j<i}(W_{ij} - W'_{ij})s_i s_j\}} \frac{e^{\{\sum_{j<i} W'_{ij} s_i s_j\}}}{Z(W')}$$

$$\equiv \left\langle \exp\left\{\sum_{j<i}(W_{ij} - W'_{ij})s_i s_j\right\} \right\rangle_{p(\mathbf{s}|W')} \quad (8)$$

where $\langle \cdot \rangle_p$ denotes expectation under $p$. Thus any method for sampling from $p(\mathbf{s}|W')$, such as MCMC methods, exact sampling methods, or any deterministic approximation that can yield the above expectation can be nested into the Metropolis sampler for $W$.

The Metropolis scheme is often not an efficient way of sampling from continuous spaces as it suffers from "random-walk" behaviour. That is, it typically takes at least order $t$ steps to travel a distance of $\sqrt{t}$. Schemes exist that use gradient information to reduce this behaviour by simulating a stochastic dynamical system [13]. The simplest of these is the "uncorrected Langevin method". Parameters are updated without any rejections according to the rule:

$$\theta'_i = \theta_i + \frac{\epsilon^2}{2} \frac{\partial}{\partial \theta_i} \log p(\mathbf{x}, \theta) + \epsilon n_i, \quad (9)$$

where $n_i$ are independent draws from a zero-mean unit variance Gaussian. Intuitively this rule performs gradient descent but explores away from the optimum through the noise term. Strictly this is only an approximation except in the limit of vanishing $\epsilon$. A corrected version would require knowing $Z(W)$ as well as the gradients. This effort may not be justified when the gradients and $Z(W)$ are only available as approximations. However approximate correction would allow use of the more general hybrid Monte Carlo method.

Using the above or other dynamical methods, a third target for approximation for systems with continuous parameters is the gradient of the joint log probability. In the case of BMs, we have:

$$\frac{\partial \log p(\mathbf{S}, W)}{\partial W_{ij}} = \sum_n s_i^{(n)} s_j^{(n)} - N \langle s_i s_j \rangle_{p(\mathbf{s}|W)} + \frac{\partial \log p(W)}{\partial W_{ij}} \quad (10)$$

Assuming an easy to differentiate prior, the main difficulty arises, as in (4), from computing the middle term: the unclamped expectations over the variables.

Interestingly, although many learning algorithms for undirected models (e.g. 4) are based on computing gradients of the form (10), and it would be simple to plug these into approximate stochastic dynamics MCMC methods to do Bayesian inference, this approach does not appear to have been investigated. We explore this approach in our experiments.

We have taken two existing sampling schemes (Metropolis and Langevin) and identified three targets for approximation to make these schemes tractable ($Z(W)$, $Z(W)/Z(W')$ and $\langle s_i s_j \rangle_{p(\mathbf{s}|W)}$). While our explicit derivations have focused on Boltzmann machines, these same expressions generalise in a straightforward way to Bayesian parameter inference in a general undirected model of the form (1). In particular, many undirected models of interest can be parameterised to have potentials in the exponential family, $\phi_j(\mathbf{x}_{C_j}, \theta_j) = \mathbf{u}_j(\mathbf{x}_{C_j})^\top \theta_j$. For such models, the key ingredient to an approximation are the expected sufficient statistics, $\langle \mathbf{u}_j(\mathbf{x}_{C_j}) \rangle$.

## 4 Approximation Schemes

Using the above concepts and focusing on Boltzmann machines we now define a variety of approximate sampling methods, by deriving approximations to one of our three target quantities in equations (7), (8) and (10).

**Naive mean field**. Using Jensen's inequality we can lower bound the log partition function as follows:

$$\log Z(W) = \log \sum_{\mathbf{s}} \exp\{\sum_{j<i} W_{ij} s_i s_j\} \\ \geq \sum_{j<i} W_{ij} \langle s_i s_j \rangle_{q(\mathbf{s})} + \mathcal{H}(q) \equiv F(W, q) \quad (11)$$

where $q(\mathbf{s})$ is any distribution over the variables, and $\mathcal{H}(q)$ is the entropy of this distribution. Defining the set of fully factorised distributions $\mathcal{Q}_{\text{mf}} = \{q : q(\mathbf{s}) = \prod_i q_i(s_i)\}$ we can find a local maximum of this lower bound $\log Z_{\text{mf}}(W) = \max_{q \in \mathcal{Q}_{\text{mf}}} F(W, q)$ using an iterative and tractable mean-field algorithm. We define the **mean-field Metropolis** algorithm as using $Z_{\text{mf}}(W)$ in place of $Z(W)$ in the acceptance probability computation (7). The expectations from the naive mean field algorithm could also be used to compute direct approximations to the gradient for use in a stochastic dynamics method (10).

**Tree-structured variational approximation**. Jensen's inequality can be used to obtain much tighter bounds than those given by the naive mean-field method. For example, constraining $q$ to be in the set of all tree-structured distributions $\mathcal{Q}_{\text{tree}}$ we can still tractably optimise the lower bound on the partition function [14], obtaining $Z_{\text{tree}}(W) \leq Z(W)$. The **tree Metropolis** algorithm is defined to use this in (7). Alternatively, expectations under the tree could also be used to form the gradient estimate for a stochastic dynamics method (10).

**Bethe approximation**. A recent justification for applying belief propagation to graphs with cycles is the relationship between this algorithm's messages and the fixed points of the Bethe free energy [15]. While this breakthrough gave a new approximation for the partition function, we are unaware of any work using it for Bayesian model selection. In the **loopy Metropolis** algorithm belief propagation is run on each proposed system, and the Bethe free energy is used to approximate the acceptance probability (7). Traditionally belief propagation is used to compute marginals; pairwise marginals can be used to compute the expectations used in gradient methods (10) or in finding partition function ratios (8). These approaches lead to different algorithms, although their approximations are clearly closely related.

**Langevin using brief sampling.** The pairwise marginals required in (9,10) can be approximated by MCMC sampling. The Gibbs sampler used in section 6.1 is a popular choice, whereas in section 6.2 a more sophisticated Swendsen-Wang sampler is employed. Unfortunately—as in maximum likelihood learning (4)—the parameter-dependent variance of these estimates can hinder convergence and introduce biases [8]. The **brief Langevin** algorithm, inspired by work on Contrastive Divergence, uses very brief sampling starting from the data, $\mathbf{S}$, which gives biased but low variance estimates of the required expectations. As the approximations in this section are run as an inner loop to the main sampler, the cheapness of brief sampling makes it an attractive option.

**Langevin using exact sampling**[2]**.** Unbiased expectations can be obtained in some systems using an exact sampling algorithm based on coupling from the past, eg [16]. Again variance could be eliminated by reuse of random numbers. This seems a promising area for future research.

**Pseudo-Likelihood.** Replacing the likelihood of the parameters with a tractable product of conditional probabilities is a common approximation in Markov random fields for image modelling. The only Bayesian approach to learning in large systems of which we are aware is in this context [17, 18]. The models used in our experiments (section 6.1) were not well approximated by the pseudo-likelihood, so we did not explore it further.

---

[2]Suggested by David MacKay.

## 5 Extension to Hidden Variables

So far we have only considered models of the form $p(\mathbf{x}|\theta)$ where all variables, $\mathbf{x}$, are observed. Often models need to cope with missing data, or have variables that are always hidden. These are often the models that would most benefit from a Bayesian approach to learning the parameters. In fully observed models in the exponential family the parameter posteriors are often relatively simple as they are log concave if the prior used is also log concave (as seen later in figure 1). The parameter posterior with hidden variables will be a linear combination of log concave functions, which need not be log concave and can be multi-modal.

In theory the extension to hidden variables is simple. First consider a model $p(\mathbf{x}, \mathbf{h}|\theta)$, where $\mathbf{h}$ are unobserved variables. The parameter posterior is still proportional to $p(\mathbf{x}|\theta)p(\theta)$, and we observe

$$p(\mathbf{x}|\theta) = \sum_{\mathbf{h}} p(\mathbf{x}, \mathbf{h}|\theta)$$

$$= \frac{1}{Z(\theta)} \sum_{\mathbf{h}} \exp\left\{\sum_j \phi_j((\mathbf{x},\mathbf{h})_{C_j}, \theta_j)\right\}$$

$$\log p(\mathbf{x}|\theta) = -\log(Z(\theta)) + \log(Z_{\mathbf{x}}(\theta)). \qquad (12)$$

That is, the sum in the second term is a partition function, $Z_{\mathbf{x}}$, for an undirected graph of the variables $\mathbf{h}$. To see this compare to (2) and consider the fixed observations $\mathbf{x}$ as parameters of the potential functions. In a system with multiple i.i.d. observations $Z_{\mathbf{x}}$ must be computed for each setting of $\mathbf{x}$. Note however that these additional partition function evaluations are for systems smaller than the original. Therefore, any method that approximates $Z(W)$ or related quantities directly from the parameters can still be used for parameter learning in systems with hidden variables.

The brief sampling and pseudo-likelihood approximations rely on settings of every variable provided by the data. For systems with hidden variables these methods could use settings from samples conditioned on the observed data. In some systems this sampling can be performed easily [8]. In section 6.2 several steps of MCMC sampling over the hidden variables are performed in order to apply our brief Langevin method.

## 6 Experiments

### 6.1 Fully observed models

Our approximate samplers were tested on three systems. The first, taken from [19], lists six binary properties detailing risk factors for coronary heart disease in 1841 men. Modelling these variables as outputs of a fully-connected Boltzmann machine, we attempted to draw samples from the distribution over the unknown weights. We can compute $Z(W)$ exactly in this system, which allows us to compare methods against a Metropolis sampler with an exact inner loop. A previous Bayesian treatment of these data also exists [10].

Many practical applications may only need a few tens of samples from the weights. We performed sampling for 100,000 iterations to obtain histograms for each of the weights (Figure 1). The mean-field, tree and loopy Metropolis methods each proposed changes to one parameter at a time using a zero-mean Gaussian with variance 0.01. The brief Langevin method used a step-size $\epsilon = 0.01$. Qualitatively the results are the same as [10], parameters deemed important have very little overlap with zero.

The mean-field Metropolis algorithm failed to converge, producing noisy and wide histograms over an ever increasing range of weights (figure 1). The sampler with the tree-based inner loop did not always converge either and when it did, its samples did not match those of the exact Metropolis algorithm very well. The loopy Metropolis and brief Langevin methods closely match the marginal distributions predicted by the exact Metropolis algorithm for most of the weights. Results are not shown for algorithms using expectations from loopy belief propagation in (10) or (8) as these gave almost identical performance to loopy Metropolis based on (7).

Our other two test systems are 100-node Boltzmann machines and demonstrate learning where exact computation of $Z(W)$ is intractable[3]. We considered two randomly generated systems, one with 204 edges and another with 500. Each of the parameters not set to zero, including the 100 biases, was drawn from a unit Gaussian. Experiments on an artificial system allow comparisons with the true weight matrix. We ensured our training data were drawn from the correct distribution with an exact sampling method [16]. This level of control would not be available on a natural data set.

The loopy Metropolis algorithm and the brief Langevin method were applied to 100 data points from each system. The model structure was provided, so that only non-zero parameters were learned. Figure 2 shows a typical histogram of parameter samples, the predictive ability over all parameters is also shown. Short runs on similar systems with stronger weights show that loopy Metropolis can be made to perform arbitrarily badly more quickly than the brief Langevin method on this class of system.

---

[3]These test sets are available online:
http://www.gatsby.ucl.ac.uk/~iam23/04blug/

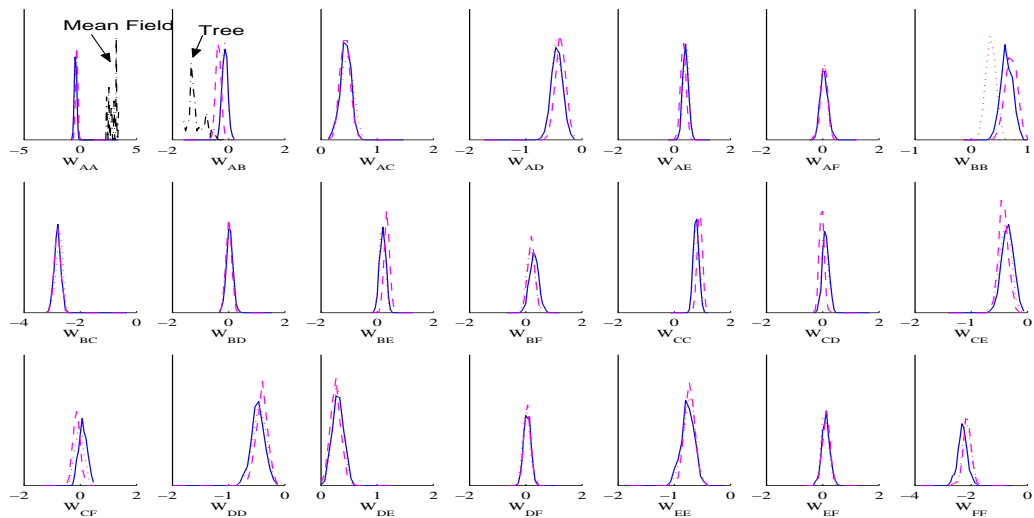

Figure 1: Histograms of samples for every parameter in the heart disease risk factor model. Results from exact Metropolis are shown in solid (blue); loopy Metropolis dashed (purple); brief Langevin dotted (red). These curves are often indistinguishable. The mean-field and tree Metropolis algorithms performed very badly; to reduce clutter these are only shown once each in the plots for $W_{AA}$ and $W_{AB}$ respectively, shown in dash-dotted (black).

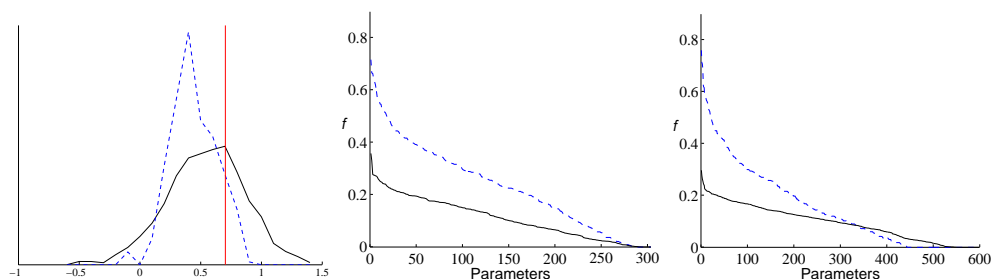

Figure 2: Loopy Metropolis is shown dashed (blue), brief Langevin solid (black). Left: an example histogram as in Figure 1 for the 204 edge BM; the vertical line shows the true weight. Also shown are the fractions of samples, $f$, within ±0.1 of the true value for every parameter in the 204 edge system (centre) and the 500 edge system (right). The parameters are sorted by $f$ for clarity. Higher curves indicate better performance.

### 6.2 Hidden variables

Finally we consider an undirected model approach taken from work on semi-supervised learning [7]. Here a graph is defined using the 2D positions, $X = \{(x_i, y_i)\}$, of unlabelled and labelled data. The variables on the graph are the class labels, $\mathbf{S} = \{s_i\}$, of the points. The joint model for the $l$ labelled points and $u$ unobserved or hidden variables is

$$p(\mathbf{S}|X, \boldsymbol{\sigma}) = \frac{1}{Z(\boldsymbol{\sigma})} \exp\left\{\sum_{i=1}^{l+u} \sum_{j<i} \delta(s_i, s_j) W_{ij}(\boldsymbol{\sigma})\right\} \quad (13)$$

where

$$W_{ij}(\boldsymbol{\sigma}) = \exp\left(-\frac{1}{2}\left(\frac{(x_i - x_j)^2}{\sigma_x^2} + \frac{(y_i - y_j)^2}{\sigma_y^2}\right)\right). \quad (14)$$

The edge weights of the model, $W_{ij}$, are functions of the Euclidean distance between points $i$ and $j$ measured with respect to scale parameters $\boldsymbol{\sigma} = (\sigma_x, \sigma_y)$. Nearby points wish to be classified in the same way, whereas far away points may be approximately uncorrelated, unless linked by a bridge of points in between.

The likelihoods in this model can be interesting functions of $\boldsymbol{\sigma}$ [7], leading to non-Gaussian and possibly multi-modal parameter posteriors with any simple prior. As the likelihood is often a very flat function over some parameter regions, the MAP parameters can change dramatically with small changes in the prior. There is also the possibility that no single settings of the parameters can capture our knowledge.

For binary classification (13) can be rewritten as a standard Boltzmann Machine. The edge weights $W_{ij}$

are now all coupled through $\boldsymbol{\sigma}$, so our sampler will only explore a two-dimensional parameter space $(\sigma_x, \sigma_y)$. However, little of the above theory is changed by this: we can still approximate the partition function and use this in a standard Metropolis scheme, or apply Langevin methods based on (10) where gradients include sums over edges.

Figure 3(a) shows an example data set for this problem. This toy data set is designed to have an interpretable posterior over $\boldsymbol{\sigma}$ and demonstrates the type of parameter uncertainty observed in real problems. We can see intuitively that we do not want $\sigma_x$ or $\sigma_y$ to be close to zero. This would disconnect all points in the graph making the likelihood small ($\approx 1/2^l$). Parameters that correlate nearby points that are the same will be much more probable under a large range of sensible priors. Neither can both $\sigma_x$ and $\sigma_y$ be large: this would force the $\times$ and $\circ$ clusters to be close, which is also undesirable. However, one of $\sigma_x$ and $\sigma_y$ can be large as long as the other stays below around one. These intuitions are closely matched by the results shown in figure 3(b). This plot shows draws from the parameter posterior using the brief Langevin method based on a Swendsen-Wang sampling inner loop described in [7]. We also reparameterised the posterior to take gradients with respect to $\log(\boldsymbol{\sigma})$ rather than $\boldsymbol{\sigma}$. This is important for any unconstrained gradient method like Langevin. Note that predictions from typical samples of $\boldsymbol{\sigma}$ will vary greatly. For example large $\sigma_x$ predicts the unlabelled cluster in the top left as mainly $\times$'s, whereas large $\sigma_y$ predicts $\circ$'s. It would not be possible to obtain the same predictive distribution over labels with a single 'optimal' setting of the parameters as was pursued in [7]. This demonstrates how Bayesian inference over the parameters of an undirected model can have a significant impact on predictions.

Figure 3(c) shows that loopy Metropolis converges to a very poor posterior distribution, which does not capture the long arms in figure 3(b). This is due to poor approximate partition functions from the inner loop. The graph induced by $W$ contains many tight cycles, which cause problems for loopy belief propagation. As expected, loopy propagation gave sensible posteriors on other problems where the observed points were less dense and formed linear chains.

## 7 Discussion

Although MCMC sampling in general undirected models is intractable, there are a variety of approximate methods that can be brought forth to tackle this problem. We have proposed and explored a range of such approximations including two variational approximations, brief sampling and the Bethe approximation, combined with Metropolis and Langevin methods. Clearly there are many more approximations that could be explored.

Note that the idea of simply constructing a joint undirected graph including both parameters and variables, and running approximate inference in this joint graph, is *not* a good idea. Marginalising out the variables in this graph results in "priors" over parameters that depend on the number of observed data points (6), which is nonsensical.

The mean field and tree-based Metropolis algorithms performed disastrously even on simple problems. We believe these failures result from the use of a lower bound as an approximation. Where the lower bound is poor, the acceptance probability for leaving that parameter setting will be exceedingly low. Thus the sampler is often attracted towards extreme regions where the bound is loose, and does not return.

The Bethe free energy based Metropolis algorithm performs considerably better and gave the best results on one of our artificial systems. However it also performed terribly on our final application. In general if an approximation performs poorly in the inner loop then we cannot expect good parameter posteriors from the outer loop. In loopy propagation it is well known that poor approximations result for frustrated systems, and systems with large weights or tight cycles.

The typically broader distributions of and less rapid failure with strong weights of brief Langevin means that we expect it to be more robust than loopy Metropolis. It gives reasonable answers on large systems where the other methods failed. We have several ideas for how to further improve upon this method, for example by reusing random seeds, which we plan to explore.

To summarise, this paper addresses the largely neglected problem of Bayesian learning in undirected models. We have described and compared a wide range of approaches to this problem, highlighting some of the difficulties and solutions. While the problem is intractable, approximate Bayesian learning should be possible in many of the applications of undirected models (**a**–**f** section 1). Examining the approximate parameter optimisation methods currently in use provides a valuable source of approximations for the quantities found in equations (7), (8) and (10). We have shown principles for using these equations to design good MCMC samplers, which should be widely applicable to Bayesian learning in these important uses of undirected models.

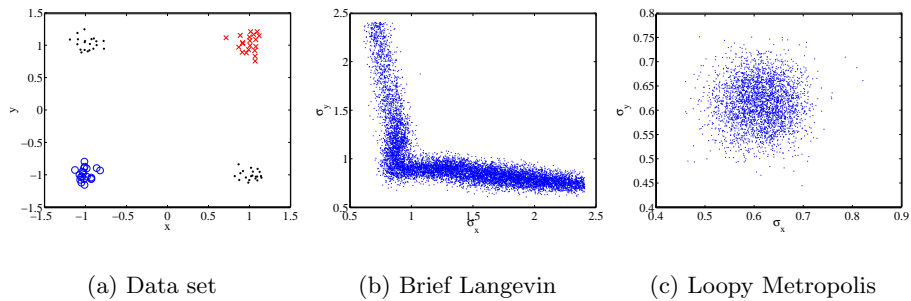

(a) Data set     (b) Brief Langevin     (c) Loopy Metropolis

Figure 3: (a) a data set for semi-supervised learning with 80 variables: two groups of classified points ($\times$ and $\circ$) and unlabelled data ($\cdot$). (b) 10,000 approximate samples from the posterior of the parameters $\sigma_x$ and $\sigma_y$ (equation 13). An uncorrected Langevin sampler using gradients with respect to $\log(\boldsymbol{\sigma})$ approximated by a Swendsen-Wang sampler was used. (c) 10,000 approximate samples using Loopy Metropolis.

## Acknowledgements

Thanks to Hyun-Chul Kim for conducting initial experiments and writing code for the mean-field and tree-based variational approximations. Thanks to David MacKay for useful comments and valuable discussions.